\documentclass[sigconf]{acmart}

\usepackage{algpseudocode}
\usepackage{algorithm}
\usepackage{amsmath}
\usepackage{graphicx}
\usepackage{subcaption}

\DeclareMathOperator{\atantwo}{atan2}

\AtBeginDocument{%
  \providecommand\BibTeX{{%
    \normalfont B\kern-0.5em{\scshape i\kern-0.25em b}\kern-0.8em\TeX}}}

\acmConference[KDD '24 (IN REVIEW)]{SIGKDD Conference on Knowledge Discovery and Data Mining}{August 25-29, 2024}{Barcelona, Spain}

\begin{document}

\title{Fingerprinting New York City's Scaffolding Problem\\ with Longitudinal Dashcam Data}

\author{Dorin Shapira}
\email{dorinruinsky@campus.technion.ac.il}
\orcid{}
\affiliation{ 
    \institution{Cornell Tech}
    \streetaddress{2 W Loop Rd}
    \city{New York}
    \state{New York}
    \country{USA}
}

\author{Matt Franchi} 
\email{mattfranchi@cs.cornell.edu}
\orcid{} 
\affiliation{ 
    \institution{Cornell Tech} 
    \streetaddress{2 W Loop Rd}
    \city{New York}
    \city{New York}
    \country{USA}
}

\author{Wendy Ju}
\email{wendyju@cornell.edu}
\affiliation{%
     \institution{Jacobs Technion-Cornell Institute, Cornell Tech}
     \state{New York}
     \city{New York}
     \country{USA}
 }

\renewcommand{\shortauthors}{Shapira et al.}

\newcommand{\numRawImages}{29,156,833\space}
\newcommand{\numScaffoldsDetected}{850,766\space}

\begin{abstract}
Scaffolds, also called sidewalk sheds, are intended to be temporary structures to protect pedestrians from construction and repair hazards. However, some sidewalk sheds are left up for years. Long-term scaffolding become eyesores, creates accessibility issues on sidewalks, and gives cover to illicit activity. 
Today, there are over 8,000 active permits for scaffolds in NYC; the more problematic scaffolds are likely expired or unpermitted. This research uses computer vision on street-level imagery to develop a longitudinal map of scaffolding throughout the city. Using a dataset of \numRawImages dashcam images taken between August 2023 and January 2024, we develop an algorithm to track the presence of scaffolding over time.  We also design and implement methods to match detected scaffolds to reported locations of active scaffolding permits, enabling the identification of sidewalk sheds without corresponding permits. We identify \numScaffoldsDetected images of scaffolding, tagging 5,156 active sidewalk sheds and estimating 529 unpermitted sheds. We discuss the implications of an in-the-wild scaffolding classifier for urban tech, innovations to governmental inspection processes, and out-of-distribution evaluations outside of New York City. 
\end{abstract}






\maketitle

\section{Introduction}
Scaffolds, or, colloquially, 'sidewalk sheds,' are temporary structures used in the construction, maintenance and repair of buildings, bridges and human-made structures. The use of scaffolds ensures the safety of workers and citizens walking near the area \cite{rubio-romero_analysis_2013}.

And yet, New York City (NYC) Mayor Eric Adams has declared a war on longstanding sidewalk sheds (\cite{heyward_mayor_2023}. Why? Because many sidewalk sheds are left up for long periods of time, even decades. NYC Local Law 11 mandates that the facades for all buildings over 6 stories high must be inspected every five years, any deficiencies must be repaired, and the public must be protected during the process. Many building owners find it easier to leave the scaffolds up than to make necessary repairs \cite{calder_city_2023}.

The prevalence of scaffolding in the urban landscape raises concerns about its impact. One issue associated with scaffolding is its detrimental impact on the urban landscape; it obstructs side views, obscures the facades of storefronts, and disrupts the overall aesthetic appeal of the surroundings \cite{deacon_planning_2013}. It creates accessibility challenges for the mobility-impaired \cite{achtemeier_impact_2019}, and helps give cover to illicit activity \cite{calder_city_2023}. 

According to NYC Open Data, there are currently more than 8,000 active permits for scaffolds in New York City \cite{nyc_opendata_dob_2024}. However, some of the scaffolds up around the city are likely unpermitted, or are left up after the permit has expired. In order for the city to 'Get Sheds Down,' as the Mayor has pledged to do, it must find ways to detect where the scaffolds are rather than just relying on official records.

In this paper, we describe a project that uses computer vision to "fingerprint," or map with fine granularity, New York City's scaffolding problem. We use longitudinal analysis of data from the networked dashcams that are installed in many of the taxis and rideshare vehicles that traverse the city's different neighborhoods. This makes it possible to profile not only not only where sidewalk sheds are now, but how long different sheds have been in place. This data medium is relatively novel (\cite{Madhavan:EECS-2017-113} and \cite{franchi_detecting_2023}), and we introduce custom methods for longitudinal analysis reliant on the temporal density afforded by crowdsourced dashcam. We see well-known hotspots of scaffolding in the Upper West Side and Upper East Side neighborhoods of Manhattan, as well as in outer neighborhoods like Brooklyn's Brownsville and Carroll Gardens. We also see new scaffolds tagged over time in our five-month dataset, as ridesharing vehicles newly make a trip to a less-visited part of the city. By training a YOLOv7 \cite{wang_yolov7_2022} object detection model to recognize scaffolds and developing a custom algorithm to confirm scaffolds with time, we establish the capability to track how long scaffolds are up in different neighborhoods, passively. We validate our findings via active sidewalk shed permit data published in the NYC Open Data by the NYC Department of Buildings \cite{nyc_opendata_dob_2024}.

The primary contribution of this work is the demonstration of crowdsourced dashcam as a medium that elevates the state-of-the-art for passive urban sensing. Vehicles that record images in our dataset aren't sent out to actively monitor sidewalk sheds; they instead gather snaps of scaffolding passively throughout each day's scores of trips. Beyond helping the municipal government and urban citizens with the challenge of infrastructure that overstays its welcome, this project highlights the way that dashcam data can make it tractable to canvas sidewalk level activity throughout the city, making it easier to address the different issues encountered in different boroughs throughout this city of over 8.5 million people.
\section{Related Work}

\subsection{Street-level image datasets}
Streetview imagery datasets, such as Google Street View, Microsoft Bing Maps, Baidu Total View, and Tencent Street View, have gained prominence in urban analytics. These datasets have been employed to conduct diverse research tasks, ranging from tree counting \cite{lu_using_2019,aikoh_comparing_2023,liu_datamix_2020}, utility pole identification \cite{zhang_using_2018}, and traffic sign detection \cite{lu_traffic_2018,balali_multi-class_2015} to bike rack \cite{maddalena_mapping_2020} and maintenance hole mapping \cite{boller_automated_2019}.
More recently, crowdsourced dashcam data has become available \cite{Madhavan:EECS-2017-113}. Here, cameras are affixed to the dashboards of ridesharing vehicles, allowing for coverage proportional to ridesharing popularity in an area. Hence, ridesharing-heavy areas like New York City have extensive coverage \cite{lam_geography_2021}. The primary advantage of crowdsourced dashcam is its much higher temporal density; GSV and similar static scene datasets offer a rich picture, but dashcam makes rich timelapses possible \cite{franchi_detecting_2023}.
Dashcam data has distinctive attributes that suit it for automated object detection in complex, detail-rich environments; it is frame-by-frame, cost-effective to obtain, variable in camera perspective, and capable of precisely capturing infrastructure, other vehicles, and small fixtures.

\subsection{Social science from street-level images}
One of the earliest studies utilizing GSV for novel social science purposes was Odgers et al.'s survey of children's neighborhoods. This work used image samples from over 1000 neighborhoods, asking coders to rate them for signs of physical disorder, physical decay, and safety. They found GSV to be a reliable and cost-effective tool for measuring both negative and positive characteristics of local neighborhoods \cite{odgers_systematic_2012}. 

Lu et al. investigated whether the streets of Hong Kong included enough greenery to encourage people to exercise, using GSV images as an analysis medium \cite{lu_using_2019}. Helbich et al. similarly looked for correlations between greenness visible in GSV and mental health in the Netherlands \cite{helbich_cant_2021}. Here, a low, static sampling density restricts longitudinal analysis and restrains specific applications. 

Gebru et al. investigated if information in GSV correlated with demographic information \cite{gebru_using_2017}. Their analysis involved the detection and classification of cars appearing in specific neighborhoods, finding that it was possible to correlate a neighborhood's vehicle makes and models with voting preferences of that neighborhood's residents. 

\subsection{Fingerprinting governmental agencies and services}
A growing number of social scientists are combining pedestrian perspectives, publicly-available datasets, and other novel data sources to address questions regarding the efficiency, productivity, and equity of governmental agencies and services. During the COVID-19 pandemic, \citet{chowdhury_tracking_2021} used dashcam data to sense social distancing compliance in New York City. In \citet{franchi_detecting_2023}, the authors use Nexar dashcam data to detect the presence of New York Police Department (NYPD) vehicles taken during 2020. Through data fusion with the publicly-accessible American Community Survey (ACS) demographics dataset, the authors uncovered significant spatial and demographic disparities over how likely different residents were likely to encounter policing vehicles. This paper formatively shows how high-frequency dashcam data, like that offered by Nexar, can be used to study governmental services. 

Prior to the above work, numerous researchers had used Google Street View (GSV), a resource offering similar visual perspectives but with a much lower sampling frequency, in pursuit of auditing city services \cite{li_assessing_2015,rundle_using_2011}.

\citet{agostini_bayesian_2023}, \citet{liu_quantifying_2023} and \citet{minkoff_nyc_2016} show that 311, New York City's governmental tool for sourcing infrastructure complaints from citizens, suffers from underreporting. Similar findings were generated outside of New York in other similar reporting systems \cite{pak_fixmystreet_2017,kontokosta_bias_2021,clark_coproduction_2013}. \citet{agostini_bayesian_2023} audits this phenomenon, and attempts to mitigate it by developing a Bayesian model that leverages a spatial correlation between street flooding reports. This work, while not utilizing image data, still shows that new computational approaches can produce more equitable government services. 

Semantically further, yet still relevant, is work that audits environmental \cite{apte_high-resolution_2017,hacker_spatiotemporal_2022}; physical \cite{hipp_measuring_2022,hara_combining_2013}; or demographic \cite{gebru_using_2017,yin_big_2015,klemmer_understanding_2021} characteristics of urban environments.

\subsection{Scaffolding inspection}
As components of the built environment ultimately meant to facilitate safety, scaffolding is subject to strict inspection standards, both at a federal level from the Occupational Health and Safety Administration (OSHA) and at a local level from the New York City Administrative Code (NYCAC). Section \text{3314.4.3} of the NYCAC allocates specific previsions for scaffolding inspection. Our analysis deals primarily with \textit{supported scaffolds}, which are constructed over sidewalks to protect pedestrians from debris that may fall from work occurring above the sidewalk \cite{nyc_department_of_buildings_code_2015}. \textit{Suspended scaffolds} are placed much higher, and hang from overhead support structures on roofs or building setbacks \cite{nyc_department_of_buildings_suspended_2023}; they generally are not visible from the fixed vantage point of a vehicular dashcam camera. For the remainder of the paper, we use 'scaffolding' for simplicity, but only consider supported scaffolds in our analysis. 

Presently, scaffolding inspection is a manual process, with audits done on downstream reports rather than the observations themselves. There are three types of inspections for supported scaffolds. Per \text{3314.4.3.3} \cite{nyc_department_of_buildings_scaffolding_2022}, installation inspections; occurring after installation of a supported scaffold, this inspection verifies that all components are in safe condition, and, if designed, are installed in accordance with the design drawings. This inspection is made by a third party acceptable to both the designer and the installer. The scaffold is not permitted to be used until the installation inspection is passed and the resulting report has been completed. 

Secondly, per \text{3314.4.3.5} \cite{nyc_department_of_buildings_scaffolding_2022}, before every construction shift or other event that might affect structural integrity, a scaffold must be inspected by a competent supervisor who can verify that all components of the scaffold remain safe to use. The results of each pre-shift report is documented daily in a report signed and dated by the individual who performed the inspection. Additionally, the scaffold cannot be used until it passes the pre-shift inspection and the resulting report has been generated. 

Lastly, per \text{3314.4.3.6} \cite{nyc_department_of_buildings_scaffolding_2022}, inspections must be made following a repair or adjustment to a scaffold at a site. The completion of these inspections blocks further work on the site. 

\subsection{Accessibility + quality of life implications}
Researchers and journalists have described the implications of widespread scaffolding for accessibility and quality of life. In the Minneapolis-St. Paul metro area, Jacob Achtemier describes a survey of nineteen normally-sighted, low vision, and blind pedestrians, stating that "construction emerged as a primary, universal theme for explaining safety challenges and
mobility efficiency with low vision and blindness," with sidewalk scaffolding being a prominent obstacle experienced \cite{achtemeier_impact_2019}. 

Leslie Deacon studied the landscape of sidewalk planning in the East Village neighborhood of Manhattan, and offers a blunt description of the sidewalk sheds permanant throughout the neighborhood; "Pedestrians must navigate these tunneled passages along entire blocks near Astor Place and other areas scattered throughout the Village, removing themselves from the typical experience of the neighborhood’s sidewalks and shifting into bland segments suffering placelessness" \cite{deacon_planning_2013}. In a 2020 New York Times article called "Our Lives, Under Construction," Dina Seiden, a Brooklyn-based author and comedian, says “People hate their fellow pedestrians in a scaffolding confinement more profoundly than they do once liberated." \cite{green_our_2020}

Overall, while seemingly understudied, existing research does find scaffolding to be a detriment to both accessibility and quality of life, and journalistic coverage is largely negative \cite{green_our_2020,yglesias_new_2022}, dubbing scaffolding a 'scourge' of the urban landscape \cite{board_shedding_2023}.

\section{Data}

\subsection{Dashcam image dataset}
\label{sec:dashcam-info}
We source a dashcam imagery dataset of NYC street scenes from Nexar, a company specializing in manufacturing cloud-connected dashboard cameras for consumer vehicles. In New York City, Nexar dashcams are often equipped on ridesharing vehicles and sometimes freight. All images are obfuscated between image capturing and later, on-premises downloading. Faces and license plates are blurred, and vehicle dashboards are cropped out via a letterboxing process. 

We assemble the dataset in-house, developing a high-speed, parallelized utility that downloads image data and metadata for each frame, and organizes data by the day of capture and h3-06 \footnote{$H3$, or Hexagonal Hierarchical Spatial Index, is a grid system developed by Uber \cite{noauthor_uberh3_2024} that segments areas of the Earth into identifiable grid cells. Nexar utilizes H3 for frame indexing, and so we work with that until merging in less arbitrary geographies like NYC boroughs, etc. The h3-06 level of the index covers a fairly large area, with an average area of $13.95$ square miles.} index.

We collect data between August 11th, 2023, and January 10th, 2024, with a period of sustained density during the entire month of November 2023. Our aim is to craft a dataset appropriate for examining both short-term changes in sidewalk sheds and longitudinal changes over longer periods. We hope that with a five-month range, we craft a dataset where at least some scaffolds are completely put up or taken down within the period of coverage.

All images are sized at 720p, 1280 pixels by width and 720 pixels by height. Aforementioned metadata includes a unique per-image frame ID, the UTC-localized timestamp when the image was taken, GPS coordinates for the capture location (accurate even in skyscraper-dense areas of Manhattan), and camera heading. Camera heading refers to the camera's orientation angle along the Z-axis in relation to global coordinates, allowing us to determine which direction a vehicle was headed at the time of image capture.

\subsection{NYC Open Data}
\label{sec:public-data}
The NYC Department of Building (DOB) maintains a comprehensive database listing all scaffolds that either currently possess or have previously obtained permit approval for their construction. This data includes essential information, such as the scaffold's coordinates, permit expiration date, permit issue date, the presence of permit renewals, and the permit type. Additional details, such as the construction firm and building obligations, although present in the database, are not within the scope of this paper.

As of January 22, 2024, there are 8,364 active sidewalk sheds in New York City, with an average age of 513 days. Scaffolds have been up for as long as 6,478 days; equivalent to over 17 years. 

\begin{figure}
    \centering
    \includegraphics[width=0.5\textwidth]{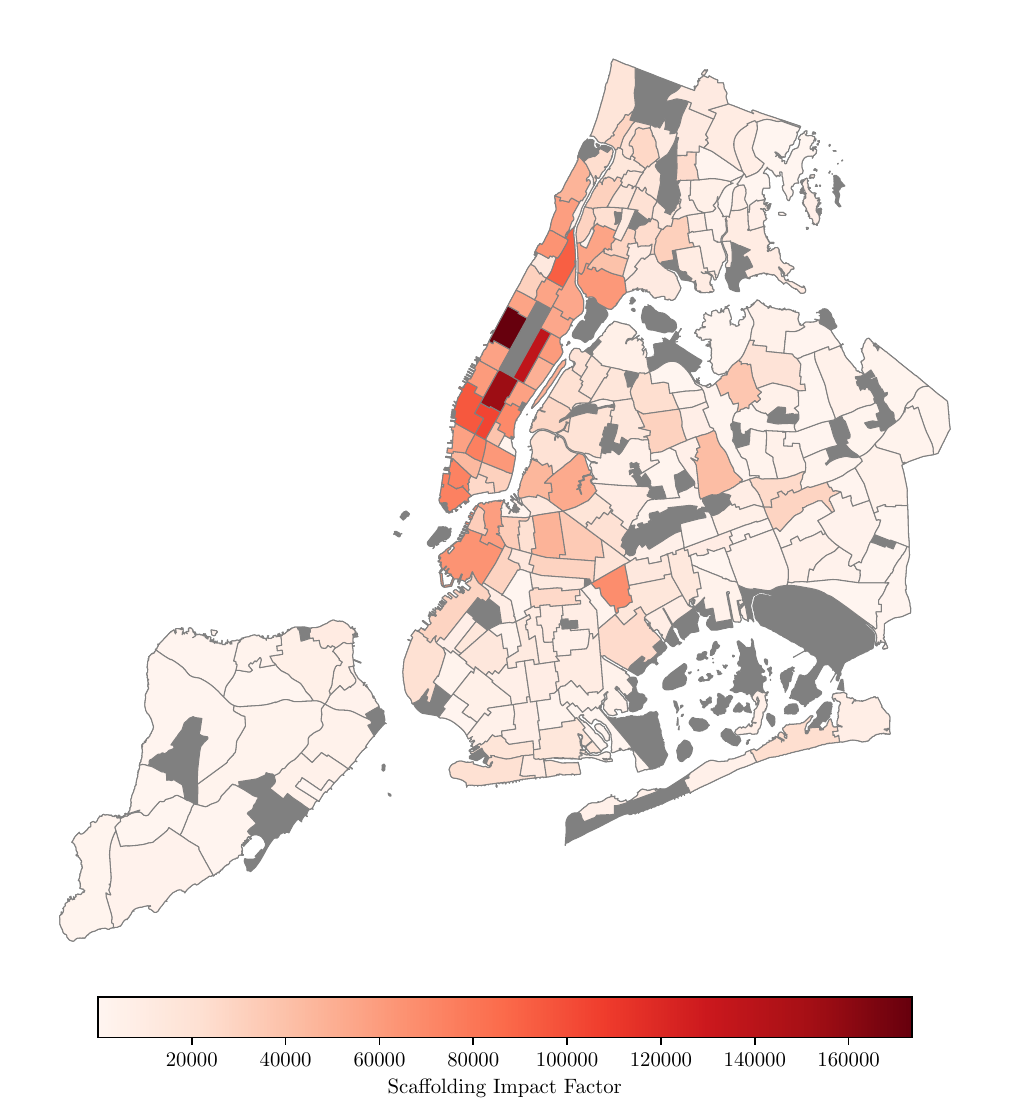}
    \caption{Citywide Distribution of Scaffolding Impact Factor, computed as the summed age of all reported scaffolds in a neighborhood $N$}
    \label{fig:scaffolding-impact}
\end{figure}

We visualize a basic 'scaffolding impact factor,' computed by taking the summed age of all scaffolds in a census area \textit{G}. For Figure \ref{fig:scaffolding-impact}, we group by Neighborhood Tabulation Area \footnote{A Neighborhood Tabulation Area (NTA) is a moderately-sized statistical geography used to report the Decennial Census and American Community Survey (ACS). They are created by the NYC Department of City Planning to report populations at a mid-size granularity, as well as for easier visual recognition.} We see scaffolding impact highest in some of Manhattan's most famous longstanding residential neighborhoods (which aligns with journalistic coverage), including the Upper West Side \cite{rag_omnipresence_2022} and Upper East Side \cite{noauthor_upper_2023}. The commercially-dominated Midtown district of Manhattan is also one the most scaffolding-dense areas of the city. Also interestingly, the Carroll Gardens and Brownsville NTA in Brooklyn also emerge as outliers in the scaffolding impact distribution. Lastly, many of the NTAs in Upper Manhattan and South Bronx have higher-than-average levels of scaffolding impact. 
 
\section{Deep learning model}
We develop a deep learning approach for in-the-wild scaffold detection in urban environments. Our methodology follows the standard supervised-learning paradigm of four steps: (1) compile and annotate a training set, (2) train a classifier on the training set, (3) compile and annotate a test set, and (4) final model evaluation on the test set. 

\subsection{Training dataset}
To put together the training dataset, we aim to identify images containing scaffolds. Unlike \cite{franchi_detecting_2023}, where the geo-spatial distribution of objects to detect was unknown, we are able to utilize public data published by the NYC Department of Buildings \cite{nyc_department_of_buildings_active_2024,nyc_opendata_dob_2024} to filter for images near reported scaffolding sites. We apply a filter to include only current active permits, denoted as having expiration dates later than 2023. We curate a dataset comprising only those images captured within a 100-meter radius of an active permit location. While we acknowledge the potential for distribution shift using this approach (compared to a truly random sample of the image distribution), we rely on the standardized appearance of scaffolding to mitigate any possible performance losses. Using Nexar metadata described in Section \ref{sec:dashcam-info} and the public datasets detailed in \ref{sec:public-data}, we calculate the proximity to the nearest permit location for each coordinate. To provide further assurance of ground-truth signal in the images, we determine whether the camera's heading was oriented toward the permit area. The pseudocode used for this section is at Algorithm \ref{alg:cap}. \\ 

\begin{figure}[h]
\centering
  \includegraphics[width=0.4\textwidth]{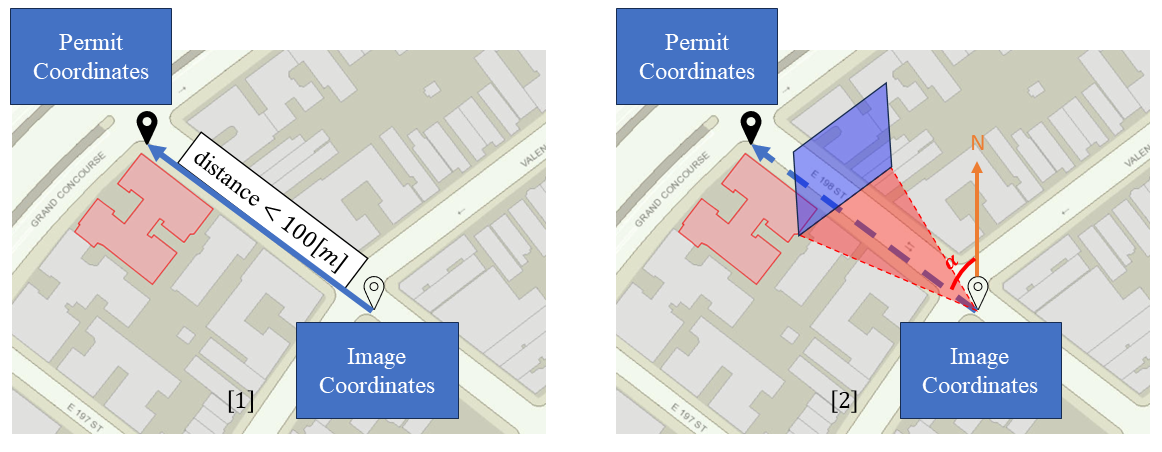}
  \caption{(1) Spatial proximity check (2) Ego Vehicle-Scaffold angle check}
 \label{fig:distance-and-angle}
\end{figure}

Figure \ref{fig:distance-and-angle} illustrates the steps for collecting the training dataset. The first step is indicated by a black pin, denoting the permit location; the building with scaffolding is also highlighted in red. The white pin marks the coordinates of the ego vehicle when the image is captured. In the first step, the left subfigure, the algorithm checks if the permit is within a 100-meter radius. In the second step, the right subfigure, $\alpha$ denotes the camera heading angle, or the camera's orientation angle relative to true north. 

To enable the trained model to learn from various scaffold representations in \textit{NYC}'s different areas, we matched the distribution of annotated data with the permit distribution across \textit{NYC} boroughs. We accomplished this by grouping images by borough and randomly selecting them from each borough, aiming to maintain a distribution similar to that of permits. The image distribution across boroughs is presented in Table \ref{tab:dist-by-borough}. Here, we compare the reported distribution of permits from NYC Open Data to the distribution of scaffolding images in our training set, stratified by NYC borough. The dissimilarity between the distributions was evaluated using KL divergence, yielding a value of 0.0038 (0.3\%).

We choose $2,214$ images for annotation. We annotate images using Label Studio, an open-source data labeling tool. We frame annotation as a rectangular object detection task. 1,015 scaffolds were annotated from those images. 

\begin{table}[H]
\centering
\begin{tabular}{ p{2cm}p{1.75cm}p{1.75cm}}
 \toprule
 Borough & Permit\newline Distribution & Dashcam \newline Distribution \\
 \hline
 Manhattan   & 0.529    &   0.538\\
 Brooklyn         &   0.220  & 0.221\\
 Bronx             &0.150 & 0.135\\
 Queens          &0.088 & 0.098\\
 Staten Island    &   0.005  & 0.011\\
 \bottomrule
\end{tabular}
\vspace{.5em}
\caption{Distribution of Permits Across NYC Boroughs}
\label{tab:dist-by-borough}
\end{table}

\subsection{Model training}
We utilized well-established models from the You Only Look Once (YOLO) suite, specifically opting for the YOLOv7-E6E object detector model. We initialized the model using pre-trained weights from the MS-COCO dataset, which encompasses 80 object categories, including people, cars, buses, traffic lights, and more.

We start with two classes. Initially, the annotations distinguished between white and green scaffolds shown in Figure~\ref{fig:types-of-scaffolds}  below, but we observed that only $7.7\%$ of the images contained white scaffolds. Due to their similar structural characteristics, this low prevalence prompted us to combine white and green scaffolds into a single scaffold class. Interesting future work might result from a stratified study, as white scaffolds are installed by the company \textit{Urban Umbrella} that specifically caters to high-wealth clientele \cite{noauthor_scaffolding_2022}.

\vspace{\baselineskip} 
\begin{figure}[h]
\begin{center}
  \includegraphics[width=0.25\textwidth]{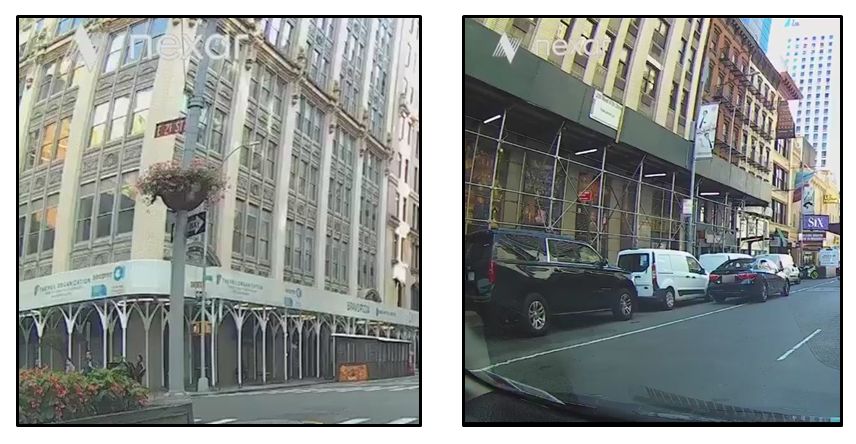}
  \caption{White and Green Scaffolds}
  \label{fig:types-of-scaffolds}
\end{center}
\end{figure}

\subsection{Model validation}

We generated our test set from random samples of Nexar images, selecting data from 15 days in August 2023. The resulting dataset, which underwent manual annotation, consisted of 2006 images. Within this dataset, we annotated 241 scaffolds.\\ The application of our trained model to this dataset produced adequate results as depicted in Figure \ref{fig:model-performance} below. Following the standard supervised learning paradigm, at the confidence threshold that maximizes the F1-Score, 0.79, we observe recall of 0.78 and precision of 0.79. 

\vspace{\baselineskip} 

\begin{figure}[h]
\begin{center}
  \includegraphics[width=0.25\textwidth]{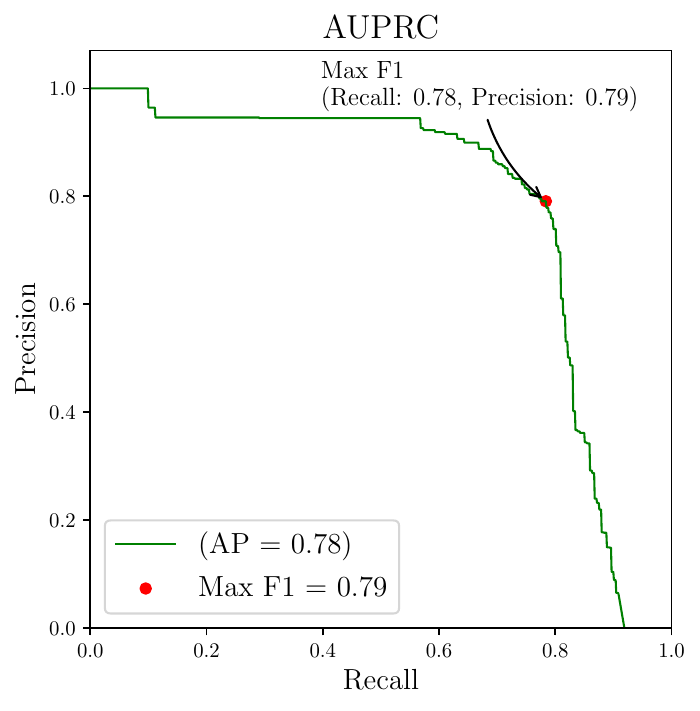}
  \caption{Precision-Recall curve for the final classifier, evaluated on the test set.}
  \label{fig:model-performance}
\end{center}
\end{figure}

\subsection{Model detections}

\begin{figure*}
     \renewcommand*{\arraystretch}{0.5}
     \centering
     \begin{subfigure}[b]{0.48\textwidth}
          \centering
         \begin{tabular}{ll}
             \includegraphics[trim={0 1.3cm 0 1.3cm}, clip, width=.5\textwidth]{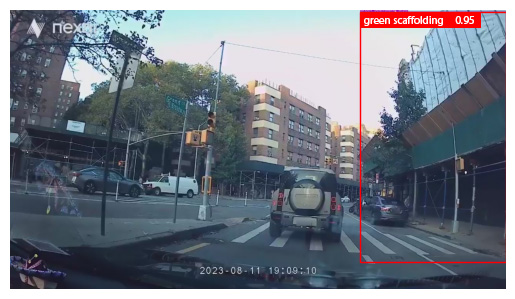} & 
             \includegraphics[trim={0 1.3cm 0 1.3cm}, clip, width=.5\textwidth]{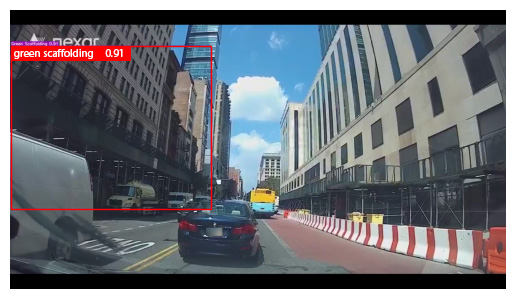} \\
             \includegraphics[trim={0 1.3cm 0 1.3cm}, clip, width=.5\textwidth]{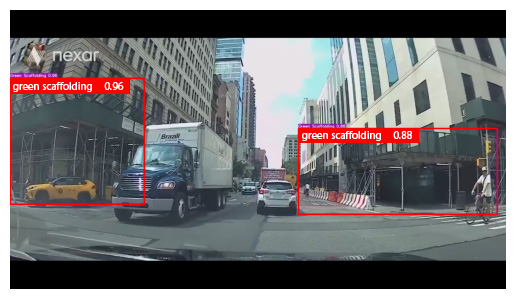} & 
             \includegraphics[trim={0 1.3cm 0 1.3cm}, clip, width=.5\textwidth]{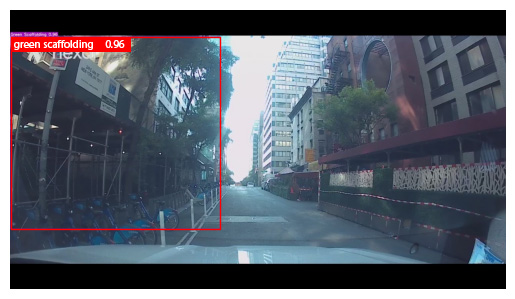} \\
          \end{tabular}
          \caption{True Positives}
          \label{fig:tp}
      \end{subfigure}
     \hfill
     \begin{subfigure}[b]{0.24\textwidth}
          \centering
          \begin{tabular}{l}
             \includegraphics[trim={0 1.3cm 0 1.3cm}, clip, width=\textwidth]{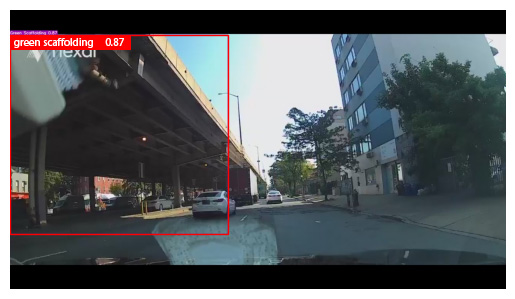} \\
             \includegraphics[trim={0 1.3cm 0 1.3cm}, clip, width=\textwidth]{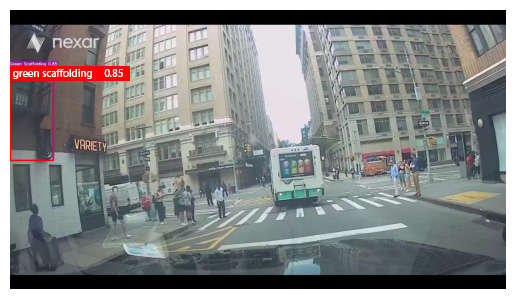} \\
          \end{tabular}
          \caption{False Positives}
          \label{fig:fp}
      \end{subfigure}
      \hfill
      \begin{subfigure}[b]{0.24\textwidth}
          \centering
          \begin{tabular}{l}
             \includegraphics[trim={0 1.3cm 0 1.3cm}, clip, width=\textwidth]{figures/final-detections/fn1_final.jpg} \\
             \includegraphics[trim={0 1.3cm 0 1.3cm}, clip, width=\textwidth]{figures/final-detections/fn2_final.jpg} \\
          \end{tabular}
          \caption{False Negatives}
          \label{fig:fn}
      \end{subfigure}
    
     \caption{Examples of scaffolding detections.} 
     \label{fig:detections}
    
 \end{figure*}

In Figure \ref{fig:detections}, we display examples of true positive detections, false positive detections of bridges and balconies as a scaffold, and false negatives, where the model misses at least one scaffold present in an image. Aware of false positives, where the model incorrectly identifies scaffolds when none exist, we aim to improve the model's performance. We move to analyze repeated detections in these areas to make an informed judgment about whether they are true positives or false negatives.

Across the period of coverage, we detect \numScaffoldsDetected scaffolds. We split these detections by month in Table \ref{tab:det-by-month}.

\begin{table}
    \begin{tabular}{ p{1.5cm}p{2cm}p{2cm}}
\toprule
Period & \# of Scaffolding Detections & \# Days of \newline Coverage \\
\midrule
Aug 2023 & 187682 & 18 \\
Sep 2023 & 84336 & 5 \\
Oct 2023 & 285127 & 13 \\
Nov 2023 & 776496 & 30 \\
Dec 2023 & 164661 & 8 \\
Jan 2024 & 8919 & 1 \\
\bottomrule
\end{tabular}

    \vspace{.5em}
    \caption{Scaffold Detections by Month/Year}
    \label{tab:det-by-month}
\end{table}

 \setlength{\tabcolsep}{0.1em} 
 \setlength{\tabcolsep}{0.5em} 

\section{Improving model accuracy}
With 78 unique days of NYC image data spanning from August 2023 to January 2024, we are enabled to develop a tagging algorithm specifically designed to scrutinize and verify detected objects. This verification process is critical to ensure the accuracy of each detection.
Given the objective of confirming scaffolds through multiple detections, the algorithm is tailored to prioritize maximum precision in the computer vision model. This focus on precision is paramount, even at the expense of recall. The rationale behind this approach is grounded in the high likelihood of accurately identifying scaffolds as true positives from repeating images of the same area. 

Our goal is to develop a highly accurate model, and while this may result in a lower recall rate, it is a calculated trade-off. Despite this emphasis on precision, it is crucial to maintain a balance and not excessively compromise recall. We have opted for a confidence threshold of 0.85 to strike this balance. This threshold is carefully chosen to enhance precision while mitigating the impact on recall as much as possible. This approach ensures that while we prioritize accuracy, we also maintain a reasonable level of recall, thereby achieving a more refined and effective algorithm. For the 85\% threshold, $r = 56.76\% $ and $p = 93.29\% $ 

\subsection{Arranging the detections}
Given the variation in detection angles and distances, a standardized method is needed to compile images showing the same scaffold. To achieve this, we implement a grid-based layout, mapping each coordinate to a grid system covering New York City. This grid is composed of rectangular 80 feet by 80 feet cells.

Our system connects with the YOLOv7 detection algorithm's output. We create a data frame for every processed image that includes the image's coordinates, capture time, and a flag indicating whether a scaffold was detected. \\
We adjust each coordinate by extending it 60 feet in the camera's direction. This adjustment is based on an analysis of various images, observing the distance between the scaffolds and the camera, which was noted to range from 0 to 120 feet.

The algorithm sorts all detections in chronological order. It then associates every row to its respective grid area, compiling a list that indicates whether a scaffold was detected. After processing the entire data frame, we accumulate a chronologically ordered list of scaffold detections or absences for each grid point.\\

The algorithm reviews each grid point by examining the last 20 detections. The number 20 is chosen so there will be enough images for the algorithm to see the area's history while balancing the algorithm's run time. From those images, we need to choose a threshold that will decide the number of detected scaffold images in the list to obtain if this area contains a scaffold.

\subsection{Confidence threshold selection}
Let $A$ denote the set of all scaffolds in NYC. Let $B$ denote the set of scaffolds detected by our classifier. Let $C$ denote the set of scaffolds tagged by the algorithm over time. Given recall $r$ and precision $p$, we define the following probabilities. $P(B|A) = r$ is the probability of the computer vision detecting a scaffold given that the scaffold exists (relevant), which is equal to 56.76\%. $P(A|B) = p$ is the probability of a scaffold being relevant given that it was detected by computer vision, which is equal to 93.29\%.

The threshold for the algorithm detection over time is denoted by $TH$, and the sum of detections $C$ from time 1 to 20 is represented by \ref{eq:C}.

\begin{equation}
    C = \sum_{i=1}^{20} B_i \geq TH
    \label{eq:C}
\end{equation}

\newcommand\eqIid{\mathrel{\stackrel{\makebox[0pt]{\mbox{\normalfont\tiny i.i.d}}}{=}}}
\newcommand\eqBin{\mathrel{\stackrel{\makebox[0pt]{\mbox{\normalfont\tiny bin.}}}{=}}}
The probability of the algorithm detecting a scaffold, given that the scaffold exists, is given by and can be derived from the equation.\\ We assume independence between the images. This is because the images are taken in different instances, with different dashcams and angles.

\begin{align}
    P(C|A) &= P\left(\sum_{i=1}^{20}B_i \geq TH |A\right) \eqIid \sum_{i=1}^{20} P(B_i|A) \geq TH \eqBin \nonumber \\
    &\eqBin \sum_{j=TH}^{20} \binom{20}{j} r^j (1-r)^{20-j}
    \label{eq:prob_of_C_knowing_A}
\end{align}

Which gives us the Recall of the new algorithm. For the probability that the scaffold exists, given the algorithm detected a scaffold:

\begin{align}
    P(A|C) &= 1 - P(\sim A|C) \nonumber \\
    &= 1 - P\left(\sim A | \sum_{i=1}^{20} B_i \geq TH \right) \nonumber \\
    &= 1 - \sum_{j=TH}^{20} \binom{20}{j} (1-p)^j (p)^{20-j}
    \label{eq:prob_of_A_knowing_C}
\end{align}

Which gives us the precision of the algorithm. Evaluating both precision and recall for each TH, we can select the threshold that gives the highest value in both. 

\begin{figure}[h]
\begin{center}
  \includegraphics[width=0.45\textwidth]{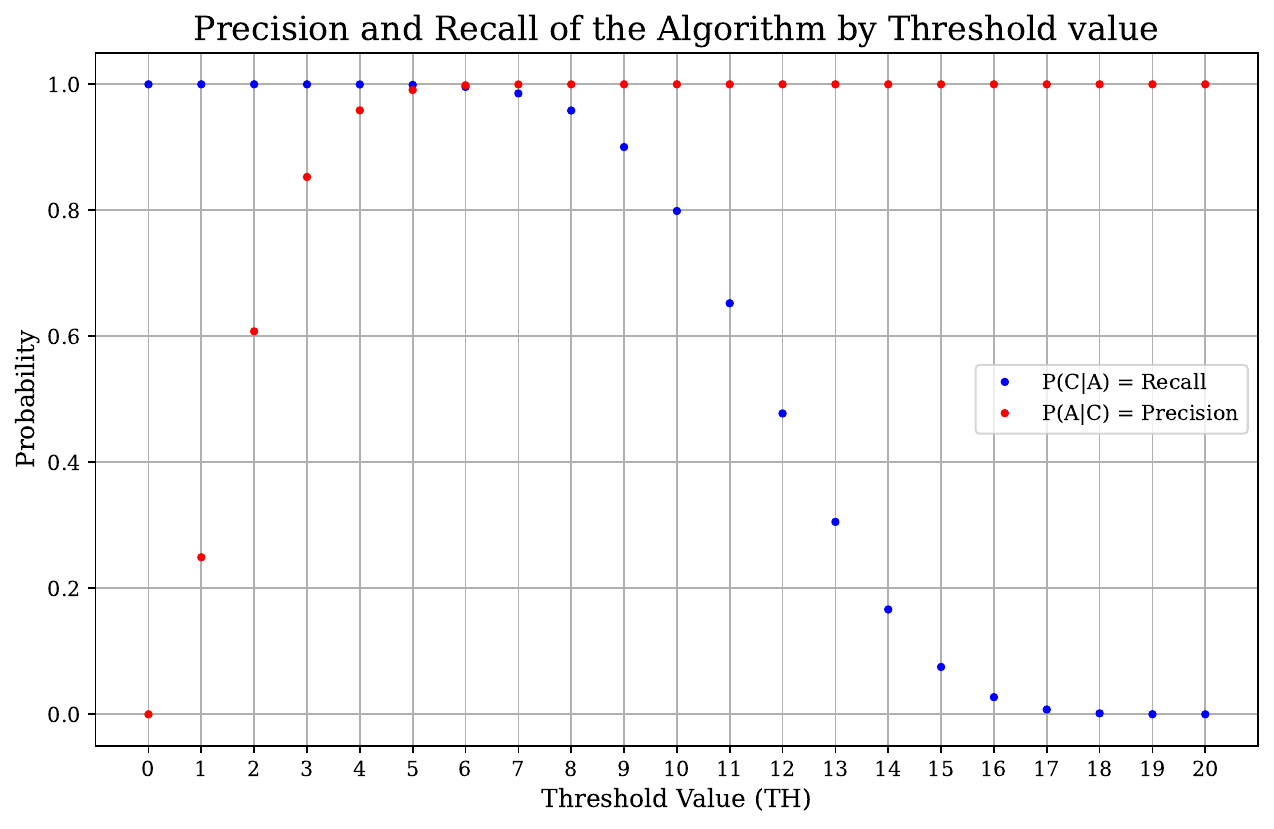}
  \label{fig: precision recall algorithm}
  \caption{Precision-Recall Curve of the tagging algorithm.}
\end{center}
\end{figure}

We evaluate TH at 5, 6, and 7. From Table \ref{tab:thresholds}, we decided on TH = 6. meaning that If at least six (or 30 \% out of the 20 detections) of these detections are positive, the area is classified as containing a scaffold with high accuracy. This analysis is applied uniformly across the grid. The tagging algorithm’s pseudocode is presented at Algorithm \ref{alg:scaffold}.

\begin{figure*}
    \includegraphics[width=\textwidth]{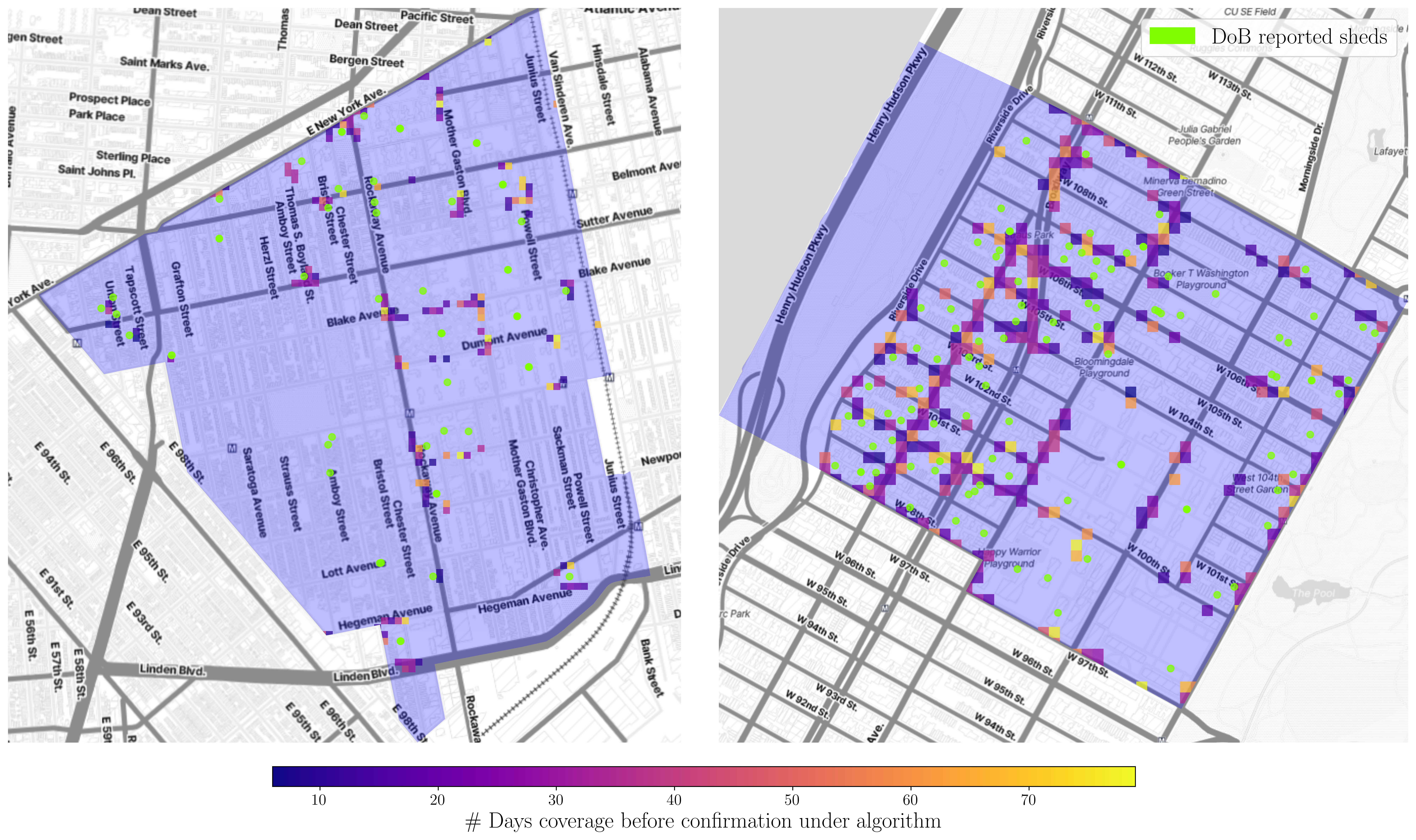}
    \caption{Zoom-in of two neighborhoods, Brownsville and the Upper West Side, showing DOB-reported scaffolds and our own detections. It is important to note that DoB-reported locations are often aligned to the building a scaffold belongs to, hence reported areas sometimes being in the interior of a block.}
    \label{fig:neighborhoods}
    
\end{figure*}

Figure \ref{fig:neighborhoods} shows the time needed to confirm different scaffolding instances in two selected neighborhoods: Brownsville, Brooklyn; and the Upper West Side, Manhattan. We also overlay reported instances of active scaffolds from the NYC Department of Buildings.

\subsection{Evaluation on DoB active shed data}
We access a dataset of active sidewalk sheds in the city as ground truth for our model detections. This evaluation is the primary way we evaluate our model's accuracy. The estimand \cite{lundberg_what_2021} is two-fold; (1) what proportion of the known active scaffolds in the city can we confirm?, (2) what proportion of known active scaffolds are out of range in our data distribution?. By combining these two evaluations, we get a ceiling of performance, and then a representation of how close we are to that ceiling. Furthermore, we also investigate how many of our detected scaffolds align with a known permit, giving us an idea of how many unpermitted scaffolds there are in the city. 

\subsubsection{Tagged scaffolds}
Tagged scaffolds are those sidewalk sheds confirmed by our tagging algorithm through longitudinal detections. To identify the number of tagged scaffolds, we align the raw coordinates from each positive image to an 80 ft x 80 ft grid system. Upon finding that this small cell size results in multiple cells per scaffold, we group cells around known active sheds together, resulting in a 320 ft x 320 ft area centered at each known coordinate. At this point, we can answer question (1) and say there are 5,685 scaffolds detected in our dashcam dataset of New York City. 

\subsubsection{Performance ceiling}
To determine a performance ceiling, we compute the number of known scaffolds that lie outside of the coverage of our dataset. We proxy the same algorithm used to tag/confirm a scaffold; so, if there are less than 20 images within 120 feet\footnote{We use 120 feet as this is the maximum distance between ego-vehicle and scaffold that we observed during annotation.} of a known sidewalk shed, we count that shed as out-of-scope. After this analysis, we find that \textit{2,512} scaffolds lie outside of scope. And so, as of January 10 2024, 2,512 out of 8,336 scaffolds are invisible in our dataset, or 30.1\%. As expected, this varies across each borough's scaffolding distribution; by borough, 19.4\% in Manhattan, 34.5\% in Brooklyn, 40.0\% in The Bronx, 43.5\% in Queens, and 74.2\% of sidewalk sheds in Staten Island are undetectable.

\subsubsection{Checking for permits}
To estimate the distribution of scaffolds without permits in our dataset, we evaluate tagged scaffolds that lie outside of any 320x320 ft area cast by an active scaffold. We find that out of our 5,685 sidewalk shed detections, 529, or 9.3\%, are unpermitted. 

\subsubsection{Summary}
We are satisfied with this model performance. Out of 8,336 known sidewalk sheds, 2,512 are undetectable due to lack of coverage, 529 are predicted to be unpermitted, and 5,156 are confirmed by our model. This means that our model missed 668 scaffolds, or 8.0\%, an unknown proportion of which are due to our coarse 320 x 320 ft detection region around each known shed.

\section{Discussion}
This analysis takes several important steps towards using large-scale, crowdsourced dashcam data for the sensing of the urban built environment. We prove it is possible to sense scaffolding both statically and longitudinally, establishing a framework to aggregate detections over time to maximize accuracy. This approach is especially well-suited for analyses that sense static or relatively immobile objects.

\subsection{Tradeoffs}
We made several design decisions with tradeoffs during this project. Firstly, some parameters of the tagging algorithm were chosen arbitrarily, like the 20-day rolling window. You would adjust this time window based on the object to be detected, and the frequency of change. Secondly, we select a confidence threshold of 0.85 for our supervised classifier, electing for higher precision and lower recall; this threshold is dependent on use case and individual model performance, and should not be blindly utilized in other projects. Lastly, we opt for a coarse, 320 ft x 320 ft detection region around each permit, which groups the too-small 80 ft x 80 ft cells together at the expense of noise in areas with highly dense scaffolding distributions like the Upper West Side neighborhood of Manhattan. 

\subsection{Expanded deployment} 
We detail several areas of future work that would strengthen the results of this project and explore our model's generalization ability. 

\subsubsection{Out-of-distribution analysis}
First, we would like to evaluate our scaffolding classifier on scaffolds from other cities around the United States, and other cities around the world. A limitation of the police vehicle classifier in \cite{franchi_detecting_2023} is that it would not generalize well, due to the unique branding placed on police vehicles in New York City. We anticipate, from visual inspection of scaffolds in other cities, that the distribution of scaffolding appearances is more homogeneous than than of police vehicles, and so classifier generalization is likely better. That said, we would still like to prove this empirically, as models that generalize well are quite valuable. 

\subsubsection{Towards efficiency gains for government}
Second, we would like to collaborate more directly with a city agency or journalistic organization in conducting further scaffolding permit compliance analysis. The problem of scaffolding has earned much news coverage throughout 2023 \cite{heyward_mayor_2023, siff___tired_2023}, encouraging public discourse. Direct collaboration with NYC's Department of Buildings might also yield the ability to pinpoint unpermitted scaffolds that should have been permitted and confirm unpermitted scaffolds that do not require permitting, for reasons like emergency work (NYCAC Section \text{28-105.4.1}) \cite{noauthor_project_nodate} and design and permit requirements (NYCBC \text{3314.2}) \cite{nyc_new_nodate}. 

Per NYCBC \text{3314.2}, "A supported scaffold that is less than 40 feet in height; not an outrigger scaffold; without any hoisting equipment over 2000 pound capacity; and is not designed to be loaded over 75 pounds per square foot, \textit{does not need a permit}." While this policy means that a wholely computational \& automated approach to scaffolding compliance detection is infeasible, at least in New York City, it remains possible to filter the backlog of manual inspection data to a quantity small enough to economically parse. 

\subsubsection{Towards on-edge, real-time detection}
Third, we would like to investigate the process of model distillation, in pursuit of real-world applications of our classifier. By distilling our model's weights into a smaller model, we could enable real-time inference on edge devices like networked dashcams \cite{cob-parro_smart_2021,murshed_machine_2022,koubaa_cloud_2022}), installed on either private or public, city-owned fleets. On-the-edge inference is not only less storage intensive, but also preserves privacy by never transferring captured images outside of the imaging device \cite{khanna_identifying_2021,liu_datamix_2020}. In this type of work, utilizing images of public street scenes and computer vision, it is imperative to consider privacy as highly as possible \cite{upmanyu_efficient_2009,xiang_being_2022,das_assisting_2017}. 

\subsection{Limitations}
We acknowledge several limitations of our work. Firstly, the grid system we used to map detections to unique scaffolding instances is a coarse 320 feet by 320 feet, and leaves room for further refinement. Secondly, our coverage of the city is limited by the distribution of ridesharing drivers in the city; as such, our ability to detect scaffolds in outer boroughs like Staten Island is diminished. Thirdly, and importantly, while we consider this a deployment of our research, it is a historical deployment. Real-time, continuous monitoring of scaffolding deployments in New York City would require much closer collaboration with both our data provider and city agencies. Lastly, and mentioned throughout, we acknowledge that this approach does not automate the scaffolding inspection process in New York City, due to complex, varied policies and requirements for manual inspections. 

\subsection{Impacts}
To our knowledge, this is the first project studying longitudinal scaffolding distributions at a city-wide scale. Not only can we assert "yes, there is a scaffold here," but we can also track scaffolds as they are constructed and taken apart over time. As mentioned earlier, this work does not automate the scaffolding inspection process required by NYC policy. However, this work would easily enable targeted inspections for agencies and city government, filtering the over 8000 scaffolds present in New York City into a manageable subset (for example, the 529 scaffolds we identify without an associated permit). Three primary groups are impacted by this work. Firstly, data scientists are presented with evidence of a promising new data medium, and are also given tools to bootstrap their own analyses. Secondly, NYC agencies are given a new way to monitor the built environment, yielding productivity and efficiency gains. Lastly, the general public is given further awareness of the extent of the city's scaffolding problem, and hopefully this work adds to the momentum of 'getting sheds down!' 

\noindent\textit{Dataset release:} As this dashcam dataset was acquired under a research evaluation license, we are restricted from publishing the raw data. However, we are in final stages of publishing a 2.4-million image dataset of NYC dashcam images, taken during 2020; this analysis would be replicable with that dataset. 

\noindent\textit{Code Release:} We will release code to replicate the analysis in this paper at this \href{https://github.com/FAR-Lab/urbanECG}{GitHub Repository}. 

\subsection{Acknowledgements}
The authors would like to thank Ilan Mandel, Maria Teresa Perreira, Emma Pierson and Eilam Shapira for helpful discussions surrounding the ideation and framing of this work. We thank Nexar for providing us with dashcam image data for this project.

\clearpage 

\bibliographystyle{stock/ACM-Reference-Format}
\bibliography{sources/matt-sources, sources/extra}
\newpage

\appendix 
\section{Supplement}

\begin{table}[ht]
\centering

\begin{tabular}{ccc}
\hline
Threshold (TH) & Recall (\%) & Precision (\%) \\ \hline
5              & 99.91       & 99.10          \\
6              & 99.59       & 99.84          \\
7              & 98.56       & 99.98          \\ \hline
\end{tabular}
\vspace{.5em}
\caption{Threshold vs. Recall and Precision}
\label{tab:thresholds}
\end{table}

\subsection{Algorithm 1}
\begin{algorithm}[h]
\caption{Finding The Nearest Permit}\label{alg:cap}
\textbf{Input:} $[X,Y]$: Permit coordinates, $[x_{i},y_{i}]$: Image coordinates, $CAM\_DEG$: Camera heading, $TH\_DIST$: Threshold distance, $TH\_DEG$: Threshold angle.\\
\textbf{Output:} List of images associated with permits for annotation.
\begin{algorithmic}[1]
\State $A = [X,Y] - [x_{i},y_{i}]$
\If{$\| A\| \leq TH\_DIST$} 
    \If{$\lvert\atantwo (A) \rvert \leq CAM\_DEG + TH\_DEG$}
        \State $Dataset \gets [x_{i},y_{i}]$
    \EndIf
\EndIf 
\end{algorithmic}
\end{algorithm}

\subsection{Algorithm 2}
\begin{algorithm}[h]
\caption{Over Time Detection}\label{alg:scaffold}
\textbf{Input:} YOLOv7 Detection, $[x_{i}, y_{i}]$: Image coordinates, $\alpha$: Camera heading, $t$: capture time. \\
\textbf{Output:} $[x^{20X20}, y^{20X20}]$ grid.
\begin{algorithmic}[1]

\For{each image in YOLOv7 output}
    \State Extract: $[x_{i}, y_{i}]$,  $t$, and detection flag
    \State  $[x'_{i}, y'_{i}] \gets [x_{i}, y_{i}] + \alpha \times 20\text{m}$
    \State Adjusted coordinates to grid  $([x^{20X20}_{i}, y^{20X20}_{i}])$
    \State  $\text{grid}([x^{20X20}_{i}, y^{20X20}_{i}]) \gets  t, l.append(\text{detection flag})$
 \EndFor
\For{$[x^{20X20}_{i}, y^{20X20}_{i}]$ in grid}
    \State $\text{last\_20} = \text{grid}([x^{20X20}_{i}, y^{20X20}_{i}]).l[:-20]$
    \If{$\text{sum}(\text{last\_20}(True))/20 \geq 30\%$}
        \State $\text{grid}([x^{20X20}_{i}, y^{20X20}_{i}]) \gets \text{'containing a scaffold'}$
    \Else
        \State $\text{grid}([x^{20X20}_{i}, y^{20X20}_{i}]) \gets \text{'not containing a scaffold'}$
    \EndIf
\EndFor
\State \textbf{return} Updated grid with scaffold detection results
\end{algorithmic}
\end{algorithm}

\end{document}